\documentclass[10pt,twocolumn,letterpaper]{article}

\usepackage{iccv}
\usepackage{times}
\usepackage{epsfig}
\usepackage{graphicx}
\usepackage{amsmath}
\usepackage{amssymb}
\usepackage{array}
\usepackage{booktabs}
\usepackage{xcolor}
\usepackage{cite}
\usepackage{multirow}
\usepackage{soul}
\usepackage{subcaption}
\usepackage{algpseudocode}
\usepackage{algorithm}
\usepackage{algorithmicx}
\usepackage{booktabs}

\usepackage{verbatim}
\usepackage{multirow, makecell}
\usepackage{lineno}
\usepackage{enumerate}
\usepackage{amsfonts}
\usepackage{gensymb}
\usepackage{bm}
\usepackage{diagbox}
\usepackage{tabularx}
\usepackage{rotating}
\newcolumntype{L}[1]{>{\raggedright\let\newline\\\arraybackslash\hspace{0pt}}m{#1}}
\newcolumntype{C}[1]{>{\centering\arraybackslash}m{#1}}
\newcolumntype{R}[1]{>{\raggedleft\let\newline\\\arraybackslash\hspace{0pt}}m{#1}}
\usepackage{xmpmulti}
\usepackage{arydshln}
\newlength\savewidth\newcommand\shline{\noalign{\global\savewidth\arrayrulewidth
  \global\arrayrulewidth 1pt}\hline\noalign{\global\arrayrulewidth\savewidth}}
\newcommand{\tablestyle}[2]{\setlength{\tabcolsep}{#1}\renewcommand{\arraystretch}{#2}\centering\footnotesize}

\usepackage[pagebackref=true,breaklinks=true,letterpaper=true,colorlinks,bookmarks=false]{hyperref}

\iccvfinalcopy % *** Uncomment this line for the final submission

\ificcvfinal\fi

\begin{document}

%%%%%%%%% TITLE
\title{OadTR: Online Action Detection with Transformers}
%
% Rethinking Online Action Detection with Transformers
%
\author{
    Xiang Wang$^1$
    \hspace{0.2cm} Shiwei Zhang$^2$
    \hspace{0.2cm} Zhiwu Qing$^1$
    \hspace{0.2cm} Yuanjie Shao$^1$
    \hspace{0.2cm} Zhengrong Zuo$^1$\\
    \hspace{0.2cm} Changxin Gao$^1$ 
    \hspace{0.2cm} Nong Sang$^1$\thanks{Corresponding Author}\\[.5ex]
    $^1$Key Laboratory of Image Processing and Intelligent Control,\\ School of Artificial Intelligence and Automation, \\Huazhong University of Science and Technology, China\\
    \hspace{0.3cm} $^2$DAMO Academy, Alibaba Group, China
    \\
    {\tt\small \{wxiang,qzw,shaoyuanjie,zhrzuo,cgao,nsang\}@hust.edu.cn, zhangjin.zsw@alibaba-inc.com}
}

%%%%%%%%% NEW COMMAND
\newcommand{\djc}[1]{{\textcolor{blue}{djc says: #1}}}
\newcommand{\mingze}[1]{{\textcolor{brown}{mingze says: #1}}}
\newcommand{\xhdr}[1]{\vspace{5pt} \noindent {\textbf{#1}}}
\definecolor{mypurple}{rgb}{0.851,0.823,0.908}
\definecolor{mygreen}{rgb}{0.812,0.878,0.823}
\definecolor{mypink}{rgb}{0.914,0.824,0.863}
\definecolor{myyellow}{rgb}{0.930,0.851,0.769}
\definecolor{mygray}{rgb}{0.745,0.745,0.745}

\maketitle
% Remove page # from the first page of camera-ready.
\ificcvfinal\thispagestyle{empty}\fi

%%%%%%%%% ABSTRACT
\begin{abstract}
Most recent approaches for online action detection tend to apply Recurrent Neural Network (RNN) to capture long-range temporal structure.
However, RNN suffers from non-parallelism and gradient vanishing, hence it is hard to be optimized.
In this paper, we propose a new encoder-decoder framework based on Transformers, named OadTR, to tackle these problems.
The encoder attached with a task token aims to capture the relationships and global interactions between historical observations. 
The decoder extracts auxiliary information by aggregating anticipated future clip representations.
Therefore, OadTR can recognize current actions by encoding historical information and predicting future context simultaneously.
We extensively evaluate the proposed OadTR on three challenging datasets: HDD, TVSeries, and THUMOS14.
The experimental results show that OadTR achieves higher training and inference speeds than current RNN based approaches, and significantly outperforms the state-of-the-art methods in terms of both mAP and mcAP. 
Code is available at \url{https://github.com/wangxiang1230/OadTR}.
%

\begin{comment}
Most recent online action detection methods adopt Recurrent Neural Network (RNN) architectures. 
%
However, RNN has the problems of non-parallelism and gradient vanishing. 
%
In the paper, from the perspective of self-attention, we propose a model that can be calculated in parallel, called \textbf{ON}line \textbf{TR}ansformer (\textbf{ONTR}). 
%
Specifically, we propose an encoder-decoder structure by using the Off-the-Shelf Transformer. 
%
Encoder attached with a task token aims to explore the relationships and global interactions between historical observations. 
%
Decoder aggregates anticipated future frames representations as auxiliary information.
%
Therefore, ONTR can not only use historical information, but also assist in identifying current actions by predicting future context.
%
We evaluate our approach on three popular online action detection datasets: HDD, TVSeries, and THUMOS14.
%
The results show that ONTR significantly outperforms the state-of-the-art methods.
%
The code will be released once this paper is accepted.
\end{comment}

\end{abstract}
%

%%%%%%%%% BODY TEXT
\section{Introduction}
The purpose of online action detection is to correctly identify ongoing actions from streaming videos without any access to the future.
Recently, this task has received increasing attention due to its great potential of diverse application prospects in real life, such as autonomous driving~\cite{self-driving}, video surveillance~\cite{surveillance}, anomaly detection~\cite{anomaly,anomaly_2}, \etc.
The crucial challenge of this task is that we need to detect the actions at the moment that video frames arrive with inadequate observations.
To solve the problem, it is important to learn the long-range temporal dependencies.

\begin{figure}[t]
\centering{\includegraphics[width=1.0\linewidth]{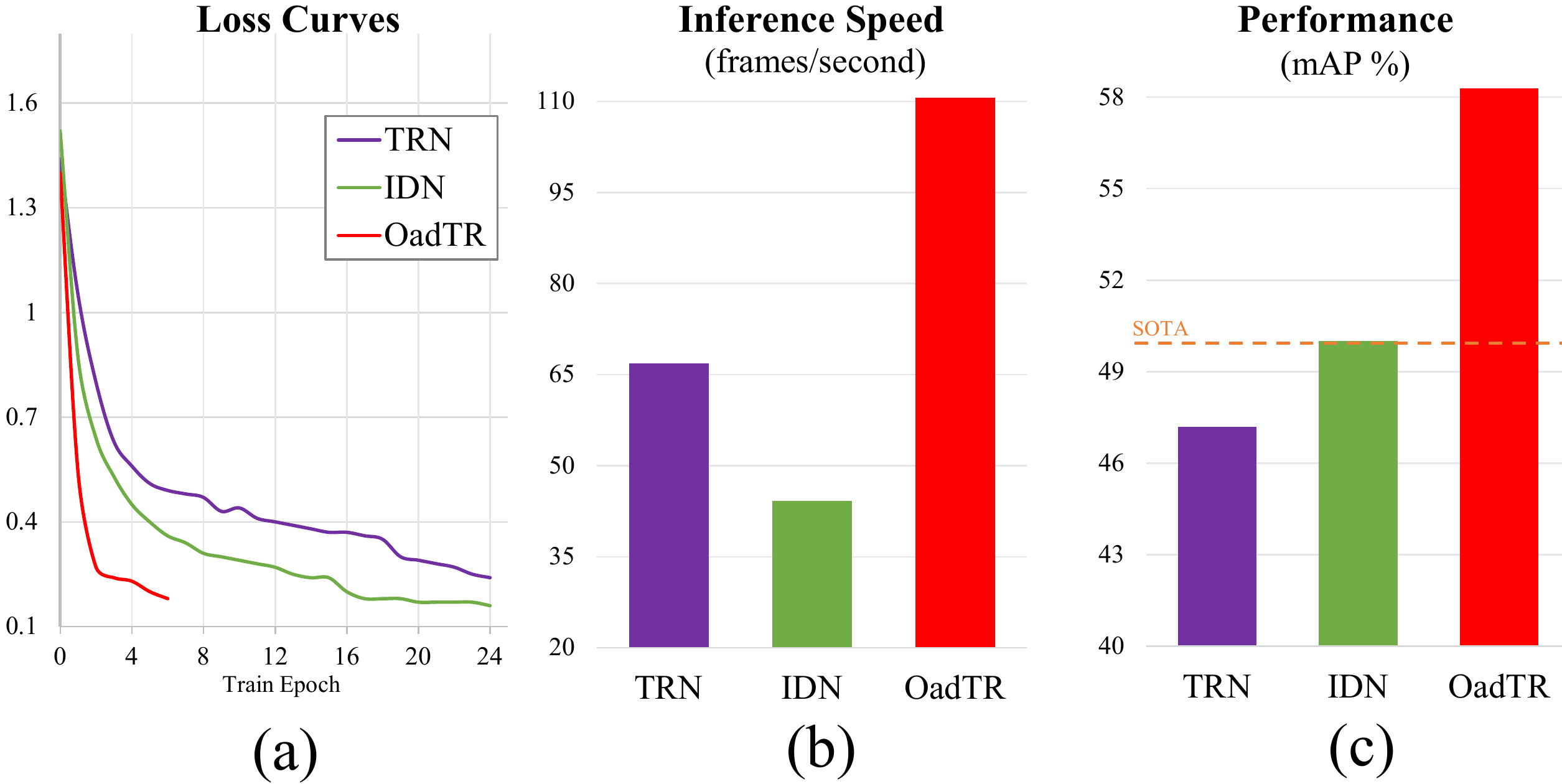}}
\vspace{-0.7cm}
\caption{\label{motivation} 
    Comparison between OadTR and the state-of-the-art online action detection methods (\ie, TRN~\cite{TRN} and IDN~\cite{IDN}): 
    (a) Comparison of the training speeds; 
    (b) Comparison of the inference speeds;
    (c) Comparison of the performance on the challenge THUMOS14 dataset.
}
\vspace{-0.4cm}
\end{figure}

Current approaches tend to apply RNN to model the temporal dependencies and have achieved impressive improvements~\cite{donahue2015long,li2016online,RED,WACV,TRN,IDN}.
% Temporal Recurrent Network (TRN)~\cite{TRN}
Typically, Information Discrimination Network (IDN)~\cite{IDN} designs a RNN like architecture to encode long-term historical information, and then conduct action recognition at current moment.
However, RNN like architectures have the problems of non-parallelism and gradient vanishing~\cite{pascanu2013difficulty,critical_RNN}. 
Thus it is hard to optimize the architectures, which may result in an unsatisfactory performance.
% was spent
This is a challenging problem for current approaches, but much less effort has been paid to solve it.
To further improve the performance, we need to design a new efficient and easily-optimized framework.
For this purpose, we propose to apply Transformers~\cite{Transformer}.
Transformers possess the strong power of long-range temporal modeling by the self-attention module, and have achieved remarkable performance in both natural language processing~\cite{Transformer,bert,Big_bird} and various vision tasks~\cite{ViT,DETR_1}.
Existing works have proved that Transformers have a better convergence than RNN architectures~\cite{transformer-vs-rnn-convergence, transformer-vs-rnn-fast}, and they are also computationally efficient.
The above properties of Transformers can naturally provide alternative scheme for the online action detection task.

\begin{figure*}[t!]
\centering{\includegraphics[width=.99\linewidth]{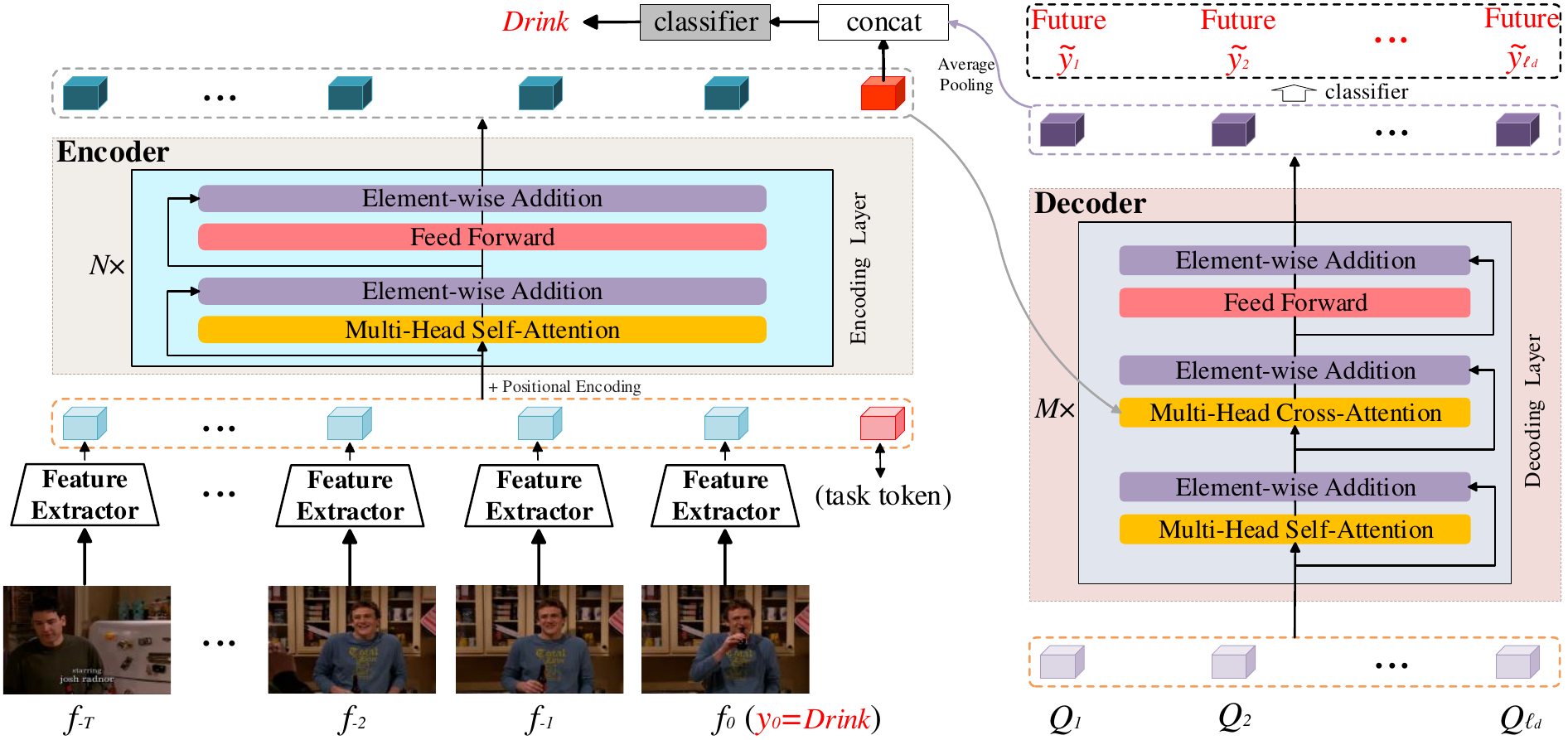}} 
\vspace{-1mm}
\caption{Illustration of the proposed Online Action Detection TRansformer (OadTR). 
Given an input streaming video $\mathbf{V}=\left \{ f_{t} \right \}_{t=-T}^{0}$, a task token is attached to the visual features output by the feature extraction network. 
Then the token feature sequence is input into the standard Transformer's encoder to model long-range historical temporal dependencies. 
Afterward, the decoder of OadTR anticipates the future context information in parallel. 
Finally, the predicted future context are involved in classifying the current action. Note that OadTR, including the encoder and decoder, is an end-to-end parallel framework. 
}
\label{network_encoder_decoder}
\vspace{-2mm}
\end{figure*}

The above observations motivate this work.
In particular, we propose a carefully designed framework, termed OadTR, by introducing the power of Transformers to the online action detection task, as illustrated in Figure~\ref{network_encoder_decoder}.
The proposed OadTR is an encoder-decoder architecture which can simultaneously learn long-range historical relationships and future information to classify current action.
The first step is to extract clip-level feature sequence from a given video by a standard CNN.
We then embed a task token to the clip-level feature sequence and input them to the encoder module.
By this means, the output of the task token can encode the global temporal relationships among the historical observations.
In contrast, the decoder is designed to predict the actions that may take place in the future moments.
Finally, we concatenate the outputs of both the task token and decoder to detect the online actions.
%
%
% In contrast, the target of the decoder is to learn future information by conducting action anticipation.
%
% Then we classify current action by aggregating both historical and future information.
%
We compare OadTR with other RNN based approaches in Figure~\ref{motivation}, which shows that the proposed OadTR is both efficient and effective.
%
% Compared with RNN based approaches, OadTR can obtain a more computational efficiency for both optimization and inference, as shown in Figure~\ref{motivation}.
%
To further demonstrate the effectiveness of OadTR, we conduct a large number of experiments on three public datasets, including HDD~\cite{HDD}, TVSeries~\cite{online_de}, and THUMOS14~\cite{thumos14}, and achieve significant improvements in terms of both mAP and mcAP metrics.

Summarily, we make the following three contributions:
\begin{itemize}

\item[$\bullet$] To the best of our knowledge, we are the first to bring Transformers into online action detection task and propose a new framework, \emph{i.e.}, OadTR;
\item[$\bullet$] We specially design the encoder and decoder of OadTR which can aggregate long-range historical information and future anticipations to improve online action detection;
\item[$\bullet$] We conduct extensive experiments, and the results demonstrate the proposed OadTR significantly outperforms state-of-the-art methods.
The massive and comprehensive ablation studies can further dissect the undergoing properties of OadTR.
\end{itemize} 

\vspace{-1pt}
\section{Related Work}
\label{sec:related_work}
\vspace{-0pt}
In this section, we will review the related methods of our approach as follows: 

\xhdr{Online Action Detection.} 
Given a live video stream, online action detection aims to identify the actions that are taking place, even if only part of the actions can be observed.
De Geest~\etal~\cite{online_de} explicitly introduced online action detection task for the first time and proposed TVSeries dataset. 
After that, they also proposed a two-stream feedback network with LSTM~\cite{LSTM} to model temporal structure~\cite{WACV}.
RED~\cite{RED} designs a Reinforced Encoder-Decoder network and a module to encourage making the right decisions as early as possible.
Several recent methods~\cite{Startnet_2,Startnet} focus on detecting action starts and minimizing the time delay of identifying the start point of an action.
IDN~\cite{IDN} directly manipulates the GRU cell~\cite{GRU} to model the relations between past information and an ongoing action.
Inspired by that humans often identify current actions by considering the future~\cite{future}, TRN~\cite{TRN} uses LSTM to predict future information recursively and combine it with past observations to identify actions. In this paper, we also introduce future information to assist in identifying current actions, but in parallel. 
% dependencies
Note that the aforementioned methods adopt RNN to model input action sequences, which are inefficient and lack of interaction between features, resulting in poor modeling capabilities for long-term dependence.

% 所有在Introduction里面出现的都应在Related Works里面出现
\xhdr{Temporal Action Detection.} 
The goal of temporal action detection is to locate the start time points and end time points of all action instances in the untrimmed video.
% MLTPN
One-stage methods~\cite{SSAD,CTCN} draw on the SSD~\cite{SSD} method in object detection and design end-to-end action detection networks with multi-layer feature pyramid structures.  
% and TAL-Net~\cite{}
Two-stage methods~\cite{RC3D, rethinking} adopt the Faster-RCNN~\cite{faster_rcnn} architecture, including proposal generation subnet and proposal classification subnet.
Most recent methods focus on generating high-quality temporal action proposals. % SSN~\cite{SSN} uses a watershed algorithm to generate action proposals.
% SSN
These methods~\cite{BSN,BMN,DBG,GTAD,BCGNN} usually locate temporal boundaries with high probabilities, then combine these boundaries as proposals and retrieve proposals by evaluating the confidence of whether a proposal contains an action within its region. 
%
%~\cite{PGCN,GTAD,BCGNN} introduce graph convolutional networks~\cite{GCN} to explore the relationships between proposals and time nodes.
%
However, the above approaches assume the entire input video can be observed, which is not available in the online task.
%

\begin{comment}
%
\xhdr{Early Action Detection.} Unlike online action detection, early action detection predicts actions from partially observed videos only containing action frames.
% This task aims to detect actions after only processing a fraction of videos.
%
Hoai~\etal~\cite{early_1} solves this problem by proposing a max-margin framework with structured SVMs. 
%
This method focus on simple scenarios, \eg, one video contains only one action.
%
Cai~\etal~\cite{early_2} employs LSTM to model temporal dependencies and designs a ranking loss for training, assuming that the gaps of predicted scores between correct and incorrect actions should be non-decreasing when a model observes more of an action.
%
\end{comment}

%In contrast to all of this existing online action detection work which only focuses on current and past observations, we 
%introduce a model that learns to simultaneously perform online action detection and anticipation of the immediate future, and 
% uses this estimated ``future information'' to improve the action detection performance of the present.
%
%
\xhdr{Transformers.} Since the success of the Transformer-based models in the natural language processing field~\cite{Transformer,bert,Big_bird}, there are many attempts to explore the feasibility of Transformers in vision tasks. 
DETR~\cite{DETR} and its variants~\cite{DETR_1,DETR_2} effectively remove the need for many hand-designed components like non-maximum suppression procedures and anchor generation by adopting Transformers. 
ViT~\cite{ViT} divides an image into $16 \times 16 $ patches and feeds them into a standard Transformer's encoder.
There are also some attempts in semantic segmentation~\cite{Semantic_1,Semantic_2}, lane shape prediction~\cite{lane}, video frame synthesis~\cite{convtransformer}, \etc. To the best of our knowledge, we are the first to introduce Transformers into the online action detection task. In particular, unlike the original auto-regressive Transformer, OadTR adopts a non-auto-regressive Transformer to generate sequences in parallel to improve efficiency. 

%
%
%--------------------------------------------Method-----------------------------------
%

%
% The Proposed Approach
\section{Methodology}

In this section, we will first introduce the problem definition and then present the proposed OadTR detailedly.

\subsection{Problem description}
%
% Online action detection task is first explicitly proposed in~\cite{online_de}.
%
Given a video stream that may contain multiple actions, the goal of the task is to identify the actions currently taking place in real-time. 
We denote $ \mathbf{V}=\left \{ f_{t} \right \}_{t=-T}^{0} $ as the input streaming video, which needs to classify the current frame chunk $f_{0}$.
We use $y_{0}$ to represent the action category of the current frame chunk $f_{0}$, and $ y_{0}\in \left \{ 0,1,...,C \right \} $, where $ C $ is the total number of the action categories and index $ 0 $  denotes the background category.
\subsection{OadTR}

To explore the potential benefits of the Transformer, we introduce the power of self-attention into the online action detection task.
In this section, we present OadTR, a encoder-decoder conformation. 
OadTR adopts an attention mechanism to capture long-range contextual information in the temporal dimension of features. 
The schematic diagram of OadTR is shown in Figure~\ref{network_encoder_decoder}. 
\subsubsection{Encoder}

\quad Given a streaming video $ V=\left \{ f_{t} \right \}_{t=-T}^{0} $, the feature extractor~\cite{TSN} extracts a 1D feature sequence by collapsing the spatial dimensions. 
Then an additional linear projection layer further maps each vectorized frame chunk feature into a $D$-dimensional feature space and $ F=\left \{ token_{t} \right \}_{t=-T}^{0} \in \mathbb{R}^{\left (T+1\right )\times D}$ denotes the resulting token sequence. 

%
%
%--------------------------------------------------------
%
\begin{figure}[t!]
\centering
\centering{\includegraphics[width=.99\linewidth]{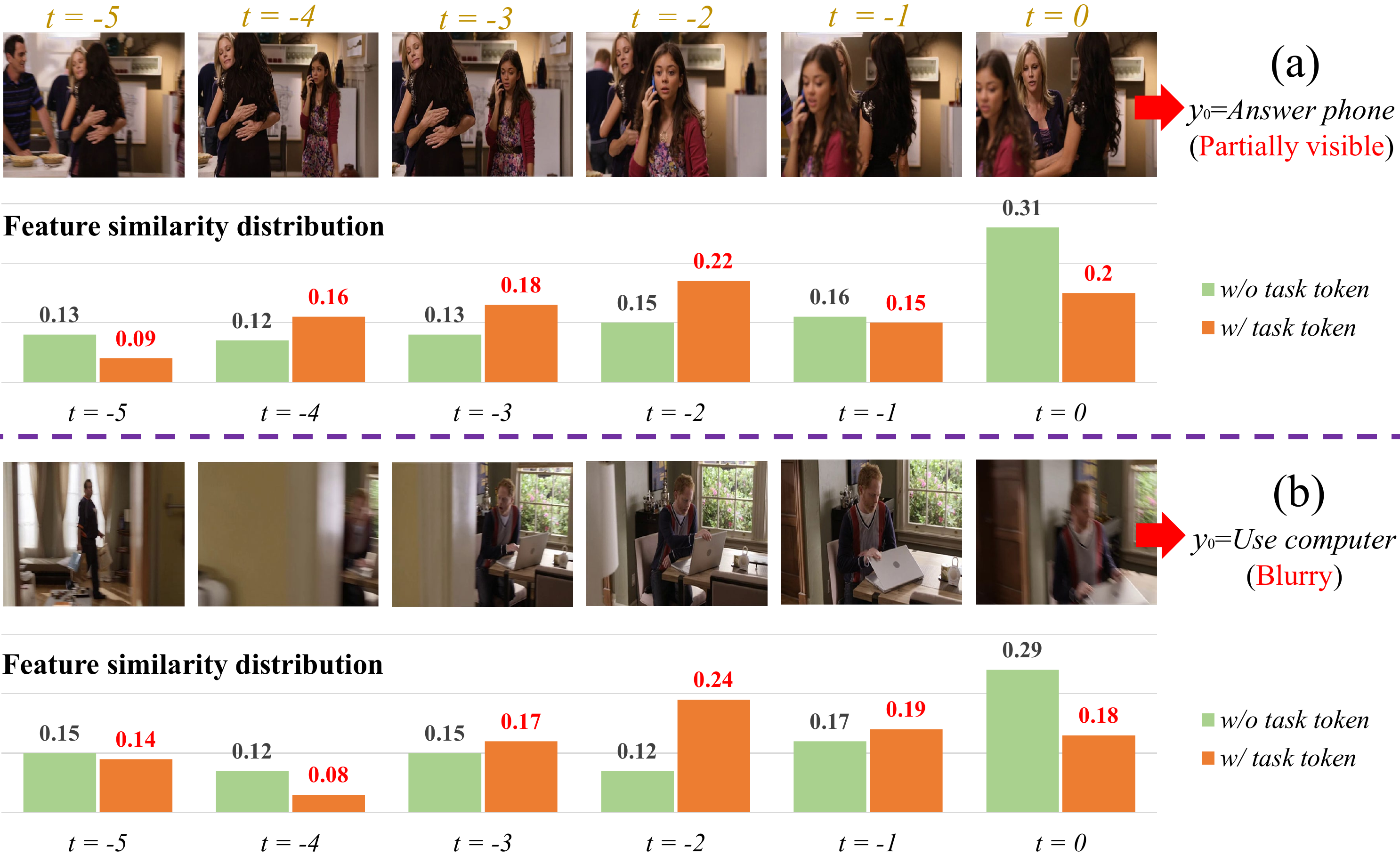}} 
% \centering{\small{(b) Information Discrimination Network (IDN)}}
%\end{minipage}
\vspace{-1mm}
\caption{\label{figure-analysis-1}Comparison of similarity distribution between the classification features (\ie, features before sending to classifier) and the input features sequence $ \tilde{F} $. Note that \textit{w/o task token}: the output classification features correspond to the token of the $f_{0}$ input; \textit{w/ task token}: the output classification features correspond to the task token. 
}
\vspace{-4mm}
\end{figure}
%
%
%
% \textbf{Task token} is first introduced in~\cite{bert}. 
In the encoder, we extend a learnable $token_{class} \in \mathbb{R}^{D} $ to the embedded feature sequence $ F $ and get the combined token feature sequence $ \tilde{F} = Stack( \left \{ token_{t} \right \}_{t=-T}^{0}, token_{class}) \in \mathbb{R}^{\left (T+2\right )\times D} $. Note that $token_{class}$ is used to learn global discriminative features related to the online action detection task. 
Intuitively, if there is no $token_{class}$ here, the final feature representation obtained by other tokens will inevitably be biased towards this specified token as a whole, and thus cannot be used to represent this learning task (\ie, \textit{w/o task token} in Figure~\ref{figure-analysis-1}). 
In contrast, the semantic embedding of $token_{class}$ can be obtained by adaptively interacting with other tokens in the encoder, which is more suitable for feature representations (\ie, \textit{w/ task token} in Figure~\ref{figure-analysis-1}). We will further confirm the necessity of $token_{class}$ in Sec~\ref{ablation}. 

Since there is no frame order information in the encoder, we need to embed position encoding additionally. Position encoding can take two forms: sinusoidal inputs and trainable embeddings. We add position encoding $E_{pos} \in \mathbb{R}^{\left (T+2\right )\times D} $ to the token sequence (\ie, element-wise addition) to retain positional information: 
\begin{equation}
X_{0} =\tilde{F}+ E_{pos}
\label{eq1}\\
\end{equation}
In this way, positional information can be kept despite the orderless self-attention. 

Multi-head self-attention (MSA) is the core component of the Transformer. Intuitively, the idea behind self-attention is that each token can interact with other tokens and can learn to gather useful semantic information more effectively, which is very suitable for capturing long-range dependencies. 
We compute the dot products of the query with all keys and apply a softmax function to obtain the weights on the values.
% respectively
Generally, the formula for self-attention is defined as: 
%\begin{equation}
%\begin{center}
\begin{align}
%X_{0} &=\tilde{F}+ E_{pos}
%\label{eq1}\\
{X}'& = {\rm Norm}(X_{0}) \\
{\rm Attention}(Q_{i}; K_{i}; V_{i}) &= {\rm softmax}\left (\frac{Q_{i}K_{i}^{T}}{\sqrt{d_{k}}}\right )V_{i}
\label{eq2}\\
H_{i} &= {\rm Attention}(Q_{i}; K_{i}; V_{i})
\label{eq3}
\end{align}
%\end{center}
%\end{equation}
where $Q_{i}={X}'\mathbf{W}_{i}^{q}$, $K_{i}={X}'\mathbf{W}_{i}^{k}$ and $V_{i}={X}'\mathbf{W}_{i}^{v}$ are linear layers applied on input sequence, and $\mathbf{W}_{i}^{q},\mathbf{W}_{i}^{k},\mathbf{W}_{i}^{v} \in \mathbb{R}^{D \times \frac{D}{N_{head}} }$. Note that queries, keys, and values are all vectors, and ${N_{head}}$ is the number of heads. 
$\frac{1}{\sqrt{d_{k}}}$ is a scaling factor, and $d_{k}$ is typically set to $D$. The scaling factor can make the training more stable and accelerate convergence. Subsequently, the outputs of heads $H_{1},H_{2},...,H_{N_{head}}$ are concatenated and feed into a linear layer. The formula is as follows:
\begin{equation}
\hat{H} = Stack\left ( H_{1},H_{2},...,H_{N_{head}} \right )\mathbf{W}_{d} \in \mathbb{R}^{\left (T+2\right ) \times D}
\label{eq4}
\end{equation}
where $\mathbf{W}_{d}$ is a linear projection. Multi-head self-attention allows the encoder to focus on multiple different patterns, which is beneficial to improve the robustness and capacity of the encoder.

Subsequently, it is followed by a two-layered feed-forward network (FFN) with GELU~\cite{GELU} activations. Meanwhile, layernorm~\cite{LN} and residual connections~\cite{resnet} are also applied. The final multiple formulas can be expressed as:
%\begin{equation}
\begin{align}
\hat{H} &={\rm MSA}(Norm(X_{0}))\\
%m_{1} &= X_{0} & \\ \Leftrightarrow
{m'}_{1} &=\hat{H} + X_{0} % \Leftrightarrow {m'}_{1}={\rm MSA}(X_{0}) 
\label{eq5}\\ 
{{m}_{n}} &={\rm FFN}\left ({\rm Norm}({m'}_{n-1})\right ) + {m'}_{n-1}
\\
{m'}_{n} &= {\rm MSA}({\rm Norm}({{m}_{n-1}})) + {{m}_{n-1}}
\label{eq6}
\end{align}
where $ n= 1,2,...,N $, $ N $ is the number of encoding layers and $ {{m}_{N}} \in \mathbb{R}^{\left (T+2\right )\times D} $ denotes the final output feature representations of the encoder.
For the convenience of explanation, we use $ {{m}_{N}^{token}}\in \mathbb{R}^D $ to signify the output representation of the encoder corresponding to the task token. 

\subsubsection{Decoder}

\quad When a person is watching a movie, he will not only remember the past, but also make predictions about what will happen in the near future~\cite{bubic2010prediction}. 
Consequently, the decoder of OadTR makes use of the observation of past information to predict the actions that will occur in the near future, so as to better learn more discriminative features. 

Prediction query $Q_{i} \in \mathbb{R}^{D'}, i=1,2,...,\ell_d$ is also learnable, where $D'$ is the number of query channels. The difference with the original Transformer~\cite{Transformer} is that our decoder decodes the $ \ell_d $ prediction queries in parallel at each decoding layer. The decoder is allowed to utilize semantic information from the encoder through the encoder-decoder cross-attention mechanism.

Here, we use $\widetilde{Q}_{i} \in \mathbb{R}^{D'}, i=1,2,...,\ell_d $ to represent the sequence output after the decoder.

\subsubsection{Training}

\quad In OadTR, we mainly use the encoder to identify the current frame chunk $f_{0}$, and the decoder to predict the coming future. At the same time, the prediction results are used as auxiliary information to better recognize the action.

For the classification task of the current frame chunk, we first concatenate the task-related features in the encoder with the pooled predicted features in the decoder. Then the resulting features go through a full connection layer and a softmax operation for action classification:
\begin{align}
\widetilde{Q} &= Avg\text{-}pool(\widetilde{Q}_{1},\widetilde{Q}_{2},...,\widetilde{Q}_{\ell_d})
\\
p_{0} &= {\rm softmax}(Concat [{m}_{N}^{token},\widetilde{Q} ]\mathbf{W}_{c})
\label{eq10}
\end{align}
where $\mathbf{W}_{c}$ represents full connection layer parameters for classification, and $p_{0} \in \mathbb{R}^{C+1}$.

In addition to the estimated current action, OadTR also outputs predicted features for the next $\ell_d$ time steps. Since future information is available during offline training, in order to ensure that a good feature expression is learned, we also conduct supervised training on future prediction features: 
\begin{equation}
\widetilde{p}^{i} = {\rm softmax}(\widetilde{Q}_{i} \mathbf{W}_{c}^{'}), i=1,2,...,\ell_d
\label{eq16}
\end{equation}
Therefore, the final joint training loss is:
\begin{equation}
Loss = {\rm CE}(p_{0},y_{0}) + \lambda \sum_{i=1}^{\ell_d}{\rm CE}(\widetilde{p}_{i}, \widetilde{y}_{i})
\label{eq14}
\end{equation}
where ${\rm CE}$ is the cross entropy loss, $ \widetilde{y}_{i} $ is the actual action category for the next step $i$ and $\lambda$ is a balance coefficient, set to 0.5 in the experiment.   
%
%
%-------------------------------------------------------------------------
%
\section{Experiments}
In this section, we evaluate the proposed OadTR on three benchmark datasets: HDD~\cite{HDD}, TVSeries~\cite{online_de}, and THUMOS14~\cite{thumos14} for online action detection. First, we compare the results between our OadTR and start-of-the-art methods. Then, we conduct more detailed ablation studies to evaluate the effectiveness of OadTR.

%-------------------------------------------------------------------------
%
%
%-------------------------------------------------------------------------
%
%
\begin{table}[t]
    \centering
    \small
\tablestyle{4pt}{1.05}
   \begin{tabular}{L{1.5cm}|C{1.5cm}|C{2.cm}|C{1.5cm}} % \hline
Method & Reference &Input & mAP (\%) \\ % \hline\hline
\shline
        {CNN}\cite{online_de}& ECCV'16 & \multirow{5}{*}{Sensors} & 22.7 \\
        {LSTM}\cite{HDD}&CVPR'18 & & 23.8 \\
        {ED}~\cite{RED}& BMVC'17 & & 27.4 \\
        {TRN}~\cite{TRN}& ICCV'19  & & 29.2 \\
        %\hline
        {\bf{OadTR}} &- & & \bf{29.8} \\
        %\hline
    \end{tabular}
    \vspace{-5pt}
    \caption{Comparison between our OadTR and other state-of-the-art online action detection methods on the HDD~\cite{HDD} dataset in terms of mAP (\%).
    }
    \vspace{-5pt}
    \label{table:HDD_detection}
\end{table}
%
%L
\begin{table}[t]
\small
\centering
\tablestyle{4pt}{1.05}
% \begin{tabular}{L{2.5cm}|C{3.cm}|C{1.5cm}}\hline 
\begin{tabular}{L{1.5cm}|C{1.5cm}|C{2.cm}|C{1.5cm}}%\hline
Method &  Reference & Input & mcAP (\%) \\ %\hline\hline
\shline
% Two-Stream
RED\cite{RED} &	BMVC'17 &\multirow{4}{*}{\shortstack{TSN-Anet}}	 & 79.2 \\
TRN\cite{TRN} & ICCV'19	&	 & 83.7 \\
IDN\cite{IDN}  & CVPR'20 &   		 & 84.7 \\
\bf{OadTR}  & -  &  		 & \bf{85.4} \\
\cdashline{1-4}
IDN\cite{IDN}  & CVPR'20 & \multirow{2}{*}{\shortstack{TSN-Kinetics}}  		 & 86.1 \\
\bf{OadTR}    & 	- &	 & \bf{87.2} \\ % \hline
\end{tabular}
\centering\caption{Comparison between our OadTR and other state-of-the-art online action detection methods on the TVSeries~\cite{online_de} dataset in terms of mcAP (\%). Note that we use the same two-stream features for fair comparison.}
\label{table:TVSeries_detection}
\end{table}
\begin{table}[t]
\small
\centering
%\begin{tabular}{C{2.5cm}|C{3.cm}|C{1.5cm}}\hline 
\tablestyle{4pt}{1.05}
\begin{tabular}{L{2.2cm}|C{1.4cm}|C{1.8cm}|C{1.1cm}}% \hline 
Method &  Reference & Setting & mAP (\%) \\ %\hline\hline  % publication
\shline
 CNN \cite{vgg}& ICLR'15 & \multirow{5}{*}{Offline} & 34.7 \\
 CNN \cite{two-stream} & NIPS'14 & & 36.2 \\
 LRCN \cite{donahue2015long} & CVPR'15 & & 39.3 \\
 MultiLSTM \cite{yeung2018every} & IJCV'18 & & 41.3 \\
 CDC \cite{CDC} & CVPR'17 & & 44.4 \\ \hline
 RED \cite{RED}& BMVC'17 & \multirow{4}{*}{\begin{tabular}[c]{@{}c@{}}Online \\ (TSN-Anet) \end{tabular}}  & 45.3 \\
%&ED \cite{gao22017bmvc} 		 & 43.7 \\
 TRN\cite{TRN} & ICCV'19  &	 & 47.2 \\
 IDN\cite{IDN}	& CVPR'20	 &      		 & 50.0 \\
 \bf{OadTR}	& -	 &      		 & \bf{58.3} \\ \cdashline{1-4}
% \multirow{4}{*}{\begin{tabular}[c]{@{}c@{}}AA\\ BB\end{tabular}}
% \multirow{4}{*}{Online (TSN-Kinetics)} & IDN\cite{IDN}      		 & 60.3 \\
 IDN\cite{IDN} & CVPR'20   &  \multirow{2}{*}{\begin{tabular}[c]{@{}c@{}}Online \\ (TSN-Kinetics) \end{tabular}} 		 & 60.3 \\
 \bf{OadTR}     & -	 & 	 & \bf{65.2} \\
% \hline
\end{tabular} 
\centering\caption{Performance comparison on the THUMOS14~\cite{thumos14} dataset in terms of mAP (\%). OadTR, IDN~\cite{IDN}, TRN~\cite{TRN}, and RED~\cite{RED} use the same two-stream features.}
\label{table:THUMOS14_detection}
\end{table}
%
%
% portion percent
%
\begin{table*}[t]
%\footnotesize
\small
\centering
\tablestyle{4pt}{1.05}
\begin{tabular}{l|cccccccccc}
%{L{3.0cm} |C{.9cm} C{.9cm} C{.9cm} C{.9cm} C{.9cm} C{.9cm} C{.9cm} C{.9cm} C{.9cm} C{.9cm}} % \hline 
%\begin{tabular}{L{2.5cm} C{1.11cm} C{1.11cm} C{1.11cm} C{1.11cm} C{1.11cm} C{1.11cm} C{1.11cm} C{1.11cm} C{1.11cm} C{1.11cm}}\hline 
%{@{\;}@{\;}l@{\;}@{\;}@{\;}@{\;}@{\;}|c@{\;}@{\;}@{\;}c@{\;}@{\;}@{\;}c@{\;}@{\;}@{\;}c@{\;}@{\;}@{\;}c@{\;}@{\;}@{\;}c@{\;}@{\;}@{\;}c@{\;}@{\;}@{\;}c@{\;}@{\,}@{\;}c@{\;}@{\,}@{\;}c@{\;}@{\;}}
& \multicolumn{10}{c}{Portion of action} \\ \cline{2-11}
Method & 0\%-10\% & 10\%-20\% & 20\%-30\% & 30\%-40\% & 40\%-50\% & 50\%-60\% & 60\%-70\% & 70\%-80\% & 80\%-90\% & 90\%-100\% \\ %\hline\hline
\shline
% \multicolumn{1}{l}{CNN \cite{online_de}} & 61.0 & 61.0 & 61.2 & 61.1 & 61.2 & 61.2 & 61.3 & 61.5 & 61.4 & 61.5 \\
\multicolumn{1}{l|}{CNN \cite{online_de}} & 61.0 & 61.0 & 61.2 & 61.1 & 61.2 & 61.2 & 61.3 & 61.5 & 61.4 & 61.5 \\
\multicolumn{1}{l|}{LSTM \cite{online_de}} & 63.3 & 64.5 & 64.5 & 64.3 & 65.0 & 64.7 & 64.4 & 64.4 & 64.4 & 64.3 \\
\multicolumn{1}{l|}{FV-SVM \cite{online_de}} & 67.0 & 68.4 & 69.9 & 71.3 & 73.0 & 74.0 & 75.0 & 75.4 & 76.5 & 76.8 \\
\multicolumn{1}{l|}{TRN \cite{TRN}} & 78.8 & 79.6 & 80.4 & 81.0 & 81.6 & 81.9 & 82.3 & 82.7 & 82.9 & 83.3 \\
\multicolumn{1}{l|}{IDN\cite{IDN}}  & \bf{80.6} & 81.1 & 81.9 & 82.3 & 82.6 & 82.8 & 82.6 & 82.9 & 83.0 & 83.9 \\
\multicolumn{1}{l|}{\bf{OadTR}}  & 79.5 & \bf{83.9} & \bf{86.4} & \bf{85.4} & \bf{86.4} & \bf{87.9} & \bf{87.3} & \bf{87.3} & \bf{85.9} & \bf{84.6} \\ % \hline
\cdashline{1-11} 
% \hline
\multicolumn{1}{l|}{IDN (Kinetics)\cite{IDN}}  & \bf{81.7} & 81.9 & 83.1 & 82.9 & 83.2 & 83.2 & 83.2 & 83.0 & 83.3 & 86.6 \\
\multicolumn{1}{l|}{ \bf{OadTR} {(Kinetics)}}  & 81.2 & \bf{84.9} & \bf{87.4} & \bf{87.7} & \bf{88.2} & \bf{89.9} & \bf{88.9} & \bf{88.8} & \bf{87.6} & \bf{86.7} \\ %\hline
\end{tabular}
\caption{Performance comparison for different portions of actions on the TVSeries dataset in terms of mcAP (\%). Note that the corresponding portions of actions are only used to compute mcAP after detecting current actions on all frame chunks in an online manner.}
\label{tab6}
\end{table*}
\subsection{Dataset and setup}
\noindent \textbf{HDD.} This dataset includes approximately 104 hours of driving actions collected in the San Francisco Bay Area, and there are a total of 11 action categories. The dataset consists of 137 sections and provides various non-visual sensors collected by the vehicle's controller area network bus. We use 100 sections for training and 37 sections for evaluation.

\noindent \textbf{TVSeries.} TVSeries contains six popular TV series, about 150 minutes for each, about 16 hours in total. The dataset totally includes 30 actions, and every action occurs at least 50 times in the dataset. TVSeries contains many unconstrained perspectives and a wide variety of backgrounds.

\noindent \textbf{THUMOS14.} This dataset has 1010 validation videos and 1574 testing videos with 20 classes. For the online action detection task, there are 200 validation videos and 213 testing videos labeled with temporal annotations. As in the previous works~\cite{TRN,IDN}, we train our model on the validation set and evaluate on the test set.

\vspace{+3pt}
\noindent \textbf{Implementation details.} For feature extractor, following previous works~\cite{RED,TRN,IDN}, we adopt the two-stream network~\cite{TSN} (3072-dimensions) pre-trained on ActivityNet v1.3~\cite{activitynet} (TSN-Anet), where spatial and temporal sub-networks adopt ResNet-200~\cite{resnet} and BN-Inception~\cite{BN} separately. For fair quantitative comparison with~\cite{IDN}, we also conduct experiments with the same TSN features (4096-dimensions) pre-trained on Kinetics~\cite{I3D} (TSN-Kinetics). 

In terms of training, we implement our proposed OadTR in PyTorch and conduct all experiments with Nvidia V100 graphics cards. Without those bells and whistles, we use Adam~\cite{adam} for optimization, the batch size is set to 128, the learning rate is set to 0.0001, and weight decay is 0.0005. Unless otherwise specified, we set $T$ to 31 for HDD dataset and 63 for both TVSeries and THUMOS14 dataset.

\vspace{+3pt}
%-------------------------------------------------------------------------
%
\noindent \textbf{Evaluation metric.} To evaluate the performance of OadTR, following the previous methods~\cite{online_de,RED,TRN,IDN}, we report per-frame mean Average Precision (mAP) on HDD and THUMOS14 dataset, and per-frame mean calibrated Average Precision~\cite{online_de} (mcAP) on TVSeries. The mAP is widely used, which needs to average the average precision (AP) of each action class. The calibrated Average Precision (cAP) can be formulated as:
\begin{align}
cPrec &= \frac{TP}{TP+\frac{FP}{w}}
\\ 
cAP &= \frac{\sum_{k}cPrec(k) \times {I}(k) }{\sum TP}
\end{align}
where ${I}(k)$ is equal to 1 if frame $ K $ is a TP. The coefficient $w$ is the ratio between negative and positive frames. 

%
%-------------------------------------------------------------------------
%
%
%
%
%##################################################################################################
\begin{table*}[t!]\centering
\hspace{1mm}
% subfloat a ---------------------------------------------------\#
\subfloat[Ablation study of the effectiveness of our proposed components on the HDD dataset.  \label{tab:HDD_compare_1}]{
\tablestyle{4pt}{1.05}
\footnotesize
%\scalebox{0.83}{
\begin{tabular}{l|c}
 %\small 
\footnotesize
\# Method & \small mAP (\%) \\
\shline
 LSTM~\cite{HDD} & 23.8 \\
 Encoder-only (Baseline) & 28.7 \\
 \hline
 Baseline + TT & 29.0 \\
 Baseline + DE & 29.2 \\
 Baseline + TT + DE (OadTR) &  \bf{29.8}\\
 %BMN &  & \underline{46.53} \\
\end{tabular}}
%}
\hspace{4mm}
% subfloat b ---------------------------------------------------
\subfloat[Ablation study of the effectiveness of our proposed components on the TVSeries dataset. \label{tab:TVSeries_compare_2}]{
\tablestyle{4pt}{1.05}
\footnotesize
%\scalebox{0.85}{
\begin{tabular}{l|c}
 %\small 
\footnotesize
\# Method & \small mcAP (\%) \\
\shline
 LSTM~\cite{IDN} & 80.9 \\
 Encoder-only (Baseline) & 84.8 \\
 \hline
 Baseline + TT & 85.0 \\
 Baseline + DE & 85.1 \\
 Baseline + TT + DE (OadTR) &  \bf{85.4}\\
\end{tabular}}
%}
\hspace{4mm}
%
% subfloat c ---------------------------------------------------
\subfloat[Ablation study of the effectiveness of our proposed components on the THUMOS14 dataset. \label{tab:THUMOS14_compare_3}]{
\tablestyle{4pt}{1.05}
\footnotesize
%\scalebox{0.82}{
\begin{tabular}{l|c}
 %\small 
\footnotesize
\# Method & \small mAP (\%) \\
\shline
 LSTM~\cite{IDN} & 46.3 \\
 Encoder-only (Baseline) & 55.8 \\
 \hline
 Baseline + TT & 56.9 \\
 Baseline + DE & 56.7 \\
 Baseline + TT + DE (OadTR) &  \bf{58.3}\\
\end{tabular}}
%}
\hspace{1mm}

\vspace{2mm}
\subfloat[Ablation study of different position encoding manners on the HDD dataset.  \label{tab:HDD_compare_1}]{
\tablestyle{4pt}{1.05}
\footnotesize
%\scalebox{0.83}{
\begin{tabular}{l|c}
 %\small 
\footnotesize
\# Position encoding & \small mAP (\%) \\
\shline
 No Position & 28.8 \\
 Fixed Position & 29.3 \\
 Learned Position &  \bf{29.8}\\
 %BMN &  & \underline{46.53} \\
\multicolumn{2}{c}{~}\\
\multicolumn{2}{c}{~}\\
\multicolumn{2}{c}{~}\\
\end{tabular}}
%}
\hspace{1mm}
% subfloat b ---------------------------------------------------
\subfloat[Ablation study of the head number on the THUMOS14 dataset. \label{tab:THUMOS14_compare_2}]{
\tablestyle{4pt}{1.05}
\footnotesize
%\scalebox{0.85}{
\begin{tabular}{l|c}
 \footnotesize
\# Head & \small mAP (\%) \\
\shline
 Head = 1 & 57.4 \\
 Head = 2 & 57.7 \\
 Head = 4 &  \bf{58.3}\\
 %BMN &  & \underline{46.53} \\
 Head = 8  & 57.7 \\
 Head = 16  & 57.8 \\
Head = 32  & 57.4 \\
\end{tabular}}
%}
\hspace{1mm}
%
% subfloat c ---------------------------------------------------
\subfloat[Ablation study of the query dimensions on the THUMOS14 dataset. \label{tab:THUMOS14_compare_3}]{
\tablestyle{4pt}{1.05}
\footnotesize
%\scalebox{0.82}{
\begin{tabular}{l|c}
 \footnotesize
\# Dim & \small mAP (\%) \\
\shline
 Dim = 128 & 56.4 \\
 Dim = 256 & 57.1 \\
 Dim = 512 & 57.3 \\
 Dim = 1024 & \bf{58.3} \\
 Dim = 1536 & 57.7 \\
 Dim = 2048 & 57.6 \\
\end{tabular}}
%}
\hspace{1mm}
%
% subfloat c ---------------------------------------------------
\subfloat[Generalization evalation. The results indicate the generalization and superiority of our model design. \label{tab:sparse_compare_3}]{
\tablestyle{4pt}{1.05}
\footnotesize
%\scalebox{0.82}{
\begin{tabular}{l|c}
 \footnotesize
\# Generalization & \small m(c)AP (\%) \\ % Spares Transformer
\shline
 Vanilla Transformer (HDD) & 29.8\\
 Sparse Transformer (HDD) & 29.6 \\
 Vanilla Transformer (TVSeries) & 85.4 \\
 Sparse Transformer (TVSeries) & 85.0 \\
 Vanilla Transformer (THUMOS14) & 58.3 \\
 Sparse Transformer (THUMOS14) & 58.1 \\
\end{tabular}}
\\
%}
%\hspace{1mm}
%
% main caption ---------------------------------------------------
\caption{Ablation studies.
\label{table:Ablation_study}}
\vspace{-4mm}
\end{table*}%\vspace{-3mm}
%
%##################################################################################################
%
%
\subsection{Compared with state-of-the-art methods}
To evaluate the performance, we compare our proposed OadTR and other state-of-the-art methods~\cite{RED,TRN,IDN} on HDD, TVSeries, and THUMOS14 datasets.
Noted that we use the same network parameter settings (\eg, $N=3$, $M=5$) when comparing different datasets. 
As illustrated in Table~\ref{table:HDD_detection}, our OadTR achieves state-of-the-art performance and improves mAP from 29.2$\%$ to 29.8$\%$ on the HDD dataset, demonstrating that our OadTR can achieve an overall performance promotion of online action detection. It could be attributed to that our OadTR introduces Transformers to obtain global historical information and the future context efficiently.

Table~\ref{table:TVSeries_detection} compares recent online action detection methods on the TVSeries dataset. To ensure a fair comparison, we adopt the same video features. The results signify that our OadTR can achieve good performance in the case of different video feature inputs. The reason for the better results of TSN-Kinetics may be that the categories of Kinetics are more diverse and contains much common generalizable presentations. 

As shown in Table~\ref{table:THUMOS14_detection}, we also conduct a comprehensive comparison on the THUMOS14 dataset. Specifically, performance is improved by 8.3$\%$ (50.0$\%$ $vs.$ 58.3$\%$ ) under TSN-Anet feature input and 4.9$\%$ (60.3$\%$ $vs.$ 65.2$\%$ ) under TSN-Kinetics feature input. The above results indicate the effectiveness of our proposed OadTR. 
When only a fraction of each action is considered, we compared OadTR with previous methods. Table~\ref{tab6} shows that our OadTR significantly outperforms state-of-the-art methods~\cite{TRN,IDN} at most time stages. Specifically, this indicates the superiority of OadTR in recognizing actions at early stages as well as all stages. It could be attributed to the ability of our OadTR to efficiently model temporal dependencies.

%
%
%-------------------------------------------------------------------------
%
\subsection{Ablation studies}
\label{ablation}
To facilitate our analysis of the model, we take the encoder without task token as our baseline.
We further conduct detailed ablation studies to evaluate different components of the proposed framework, include the following:\\
\textbf{\textit{Encoder-only (Baseline)}:} We adopt the encoder in the original Transformer~\cite{Transformer} and apply it directly to the online action detection task. Note that compared with our OadTR's encoder, the task token is missing, and the classifier is applied to the last output representation (\ie, corresponding to $f_{0}$) of the Transformer's encoder.
\\
\textbf{\textit{Baseline + TT}:} We incorporate \textit{Baseline} and the task token (TT) together, which is our OadTR's encoder. This method adds a task-related token, and we use ablation experiments to illustrate the necessity of this token. \\
\textbf{\textit{Baseline + DE}:} In this method, we add the decoder (DE) of the prediction task in OadTR to the baseline method to test and verify the function of the decoder. \\
\textbf{\textit{Baseline + TT + DE (OadTR)}:} This is the method we propose in this paper, adding task token and decoder to the baseline method together. 

In Table~\ref{table:Ablation_study} (a-c), we report the performance comparison experiments of the above methods on HDD, TVSeries, and THUMOS14 datasets. The results of \textit{Baseline + TT} demonstrate that using the additional task token is helpful for action classification. The results of \textit{Baseline + DE} explain the remarkable power of the auxiliary prediction task. When combined with the above two improvements, our OadTR (\ie, \textit{Baseline + TT + DE}) achieves the best results on three datasets. Specifically, when compared with the baseline method, our OadTR approach has improved by 1.1$\%$, 0.6$\%$, and 2.5$\%$ on HDD, TVSeries, and THUMOS14 datasets, respectively. 
\vspace{+3pt} 

% 
%
%
% predict ablation
%
\begin{table}[t]
    \centering
    \footnotesize
\tablestyle{4pt}{1.05}
% \small
    \begin{tabular}
        {@{\;\;}l@{\;\;}@{\;\;}|l@{\;\;}@{\;\;}c@{\;\;}@{\;\;}c@{\;\;}@{\;\;}c@{\;\;}@{\;\;}c@{\;\;}}
        %\toprule
        & & \multicolumn{4}{c}{Decoder steps ($\ell_d$)}\\
        % \cmidrule(r){3-6}  % {2\cmidrulekern}
\cline{3-6}
        Dataset & Task & 2 & 4 & 8 & 16 \\
        % \midrule
\shline
%\hline
        \multirow{2}{*}{HDD}
        & {Online Action Detection}   & 28.3 & 28.4 & \bf{29.8} & 28.8 \\
        & {Action Anticipation}       & 27.2 & 26.2 & 23.0 & 16.9 \\ 
        %\midrule
\hline
        \multirow{2}{*}{TVSeries}
        & {Online Action Detection}   & 85.3 & 85.2 & \bf{85.4} & 85.2 \\
        & {Action Anticipation}       & 81.8 & 80.4 & 77.8 & 74.3 \\ 
        %\midrule
\hline
        \multirow{2}{*}{THUMOS14}
        & {Online Action Detection}   & 57.4 & 57.8 & \bf{58.3} & 58.0 \\
        & {Action Anticipation}       & 53.5 & 51.0 & 45.9 & 40.7 \\
        %\bottomrule

    \end{tabular}
    \vspace{-5pt}
    \caption{
        Online action detection and action anticipation results of our proposed OadTR
        with decoder steps $\ell_d=2,4,8,16$.
    }
    \vspace{-8pt}
    \label{table:decoder}
\end{table}
%
%
% encoder layers ablation
%
\begin{table}[t]
    \centering
    \footnotesize
    \begin{subtable}[t]{\linewidth}
\centering
\tablestyle{4pt}{1.05}
    \begin{tabular}
        {@{\;\;}l@{\;\;}@{\;\;}|l@{\;\;}@{\;\;}c@{\;\;}@{\;\;}c@{\;\;}@{\;\;}c@{\;\;}@{\;\;}c@{\;\;}@{\;\;}c@{\;\;}@{\;\;}c}
        %\toprule
        & & \multicolumn{6}{c}{Encoding layers ($N,M=5$)}\\
        %\cmidrule(r){3-8}  % {2\cmidrulekern}
\cline{3-8}
        Method & Dataset & 1 & 2 & 3 & 4 & 5 & 6 \\
        %\midrule
\shline
        \multirow{3}{*}{OadTR}
        & {HDD}   & 28.8 & 29.0 & \bf{29.8} & 29.5 & 28.4 & 28.4 \\ %\cmidrule(r){2-8}
        & {TVSeries}   & 85.4 & \bf{85.5} & 85.4 & 85.3 & 84.8 & 85.0 \\ %\cmidrule(r){2-8}
        & {THUMOS14}   & 57.5 & 56.9 & \bf{58.3} & 57.4 & 56.9 & 56.6 \\
        %\bottomrule

    \end{tabular}
    \vspace{-3pt}
    \caption{
        Online action detection results of different encoding layers. Note that we fixed $M=5$ for simplicity. 
    }
    \vspace{+3pt}
\end{subtable}
\begin{subtable}[t]{\linewidth}
        \centering
\tablestyle{4pt}{1.05}
\begin{tabular}
        {@{\;\;}l@{\;\;}@{\;\;}|l@{\;\;}@{\;\;}c@{\;\;}@{\;\;}c@{\;\;}@{\;\;}c@{\;\;}@{\;\;}c@{\;\;}@{\;\;}c@{\;\;}@{\;\;}c}
        %\toprule
        & & \multicolumn{6}{c}{Decoding layers ($M, N=3$)}\\
        % \cmidrule(r){3-8}  % {2\cmidrulekern}
\cline{3-8}
        Method & Dataset & 1 & 2 & 3 & 4 & 5 & 6 \\
        %\midrule
\shline
        \multirow{3}{*}{OadTR}
        & {HDD}   & 27.7 & 28.0 & 29.5 & 29.5 & \bf{29.8} & 28.6 \\ %\cmidrule(r){2-8}
        & {TVSeries}   & 84.8 & 84.8 & 85.2 & 85.4 & 85.4 & \bf{85.5} \\ %\cmidrule(r){2-8}
        & {THUMOS14}   & 57.4 & 57.9 & 56.9 & 57.2 & \bf{58.3} & 56.6 \\
        %\bottomrule

    \end{tabular}
    \vspace{-3pt}
    \caption{
        Online action detection results of different decoding layers. Note that we fixed $N=3$ for simplicity. 
    }
    \vspace{-5pt}
\end{subtable}
\caption{
        Ablation study of encoding layers $N$ and decoding layers $M$ using TSN-Anet features.
    }
    \label{table:encoder_decoder}
\vspace{-8pt}
\end{table}
\begin{table*}[t!]
    \small % small \footnotesize
    \centering
\tablestyle{4pt}{1.05}
    \begin{subtable}[t]{\textwidth} % \textwidth  \linewidth
        \centering
        \begin{tabular}
            {@{\;}@{\;}l@{\;}@{\;}@{\;}@{\;}@{\;}@{\;}@{\;}@{\;}|c@{\;}@{\;}@{\;}@{\;}@{\;}@{\;}@{\;}@{\;}c@{\;}@{\;}@{\;}@{\;}@{\;}@{\;}@{\;}@{\;}c@{\;}@{\;}@{\;}@{\;}@{\;}@{\;}@{\;}@{\;}c@{\;}@{\;}@{\;}@{\;}@{\;}@{\;}@{\;}@{\;}c@{\;}@{\;}@{\;}@{\;}@{\;}@{\;}@{\;}@{\;}c@{\;}@{\;}@{\;}@{\;}@{\;}@{\;}@{\;}@{\;}c@{\;}@{\;}@{\;}@{\;}@{\;}@{\;}@{\;}@{\;}c@{\;}@{\,}@{\;}r@{\;\;\;}@{\;}@{\;}}
            %\toprule
            & \multicolumn{8}{c}{Time predicted into the future (seconds)} & \\
            %\cmidrule(r){2-9}  
\cline{2-9}    
Method & 0.25s & 0.5s & 0.75s & 1.0s & 1.25s & 1.5s & 1.75s & 2.0s &\;\;\;\; \;\;\;\;Avg \\
            %\midrule \midrule
\shline
            {ED~\cite{RED}} & 78.5 & 78.0 & 76.3 & 74.6 & 73.7 & 72.7 & 71.7 & 71.0 & 74.5 \\
            {RED~\cite{RED}} & 79.2 & 78.7 & 77.1 & 75.5 & 74.2 & 73.0 & 72.0 & 71.2 & 75.1 \\
            {TRN~\cite{TRN}} & 79.9 & 78.4 & 77.1 & 75.9 & 74.9 & 73.9 & 73.0 & 72.3 & 75.7 \\
            {\bf{OadTR}} & 81.9 & 80.6 & 79.4 & 78.2 & 77.1 & 76.0 & 75.2 & 74.3 & \textbf{77.8} \\ \cdashline{1-10}
            {\bf{OadTR (Kinetics)}}& 84.1 & 82.6 & 81.3 & 80.1 & 78.9 & 77.7 & 76.7 & 75.7 & \textbf{79.1} \\
            % \bottomrule
        \end{tabular}
        \vspace{-3pt}
        \caption{
            Results on the TVSeries dataset in terms of mcAP (\%).
        }
        \vspace{3pt}
    \end{subtable}
\tablestyle{4pt}{1.05}
    \begin{subtable}[t]{\textwidth}
        \centering
        \begin{tabular}
            {@{\;}@{\;}l@{\;}@{\;}@{\;}@{\;}@{\;}@{\;}@{\;}@{\;}|c@{\;}@{\;}@{\;}@{\;}@{\;}@{\;}@{\;}@{\;}c@{\;}@{\;}@{\;}@{\;}@{\;}@{\;}@{\;}@{\;}c@{\;}@{\;}@{\;}@{\;}@{\;}@{\;}@{\;}@{\;}c@{\;}@{\;}@{\;}@{\;}@{\;}@{\;}@{\;}@{\;}c@{\;}@{\;}@{\;}@{\;}@{\;}@{\;}@{\;}@{\;}c@{\;}@{\;}@{\;}@{\;}@{\;}@{\;}@{\;}@{\;}c@{\;}@{\;}@{\;}@{\;}@{\;}@{\;}@{\;}@{\;}c@{\;}@{\,}@{\;}r@{\;\;\;}@{\;}@{\;}}
            % \toprule
            & \multicolumn{8}{c}{Time predicted into the future (seconds)} & \\
            %\cmidrule(r){2-9}    
\cline{2-9}        
Method & 0.25s & 0.5s & 0.75s & 1.0s & 1.25s & 1.5s & 1.75s & 2.0s &\;\;\;\; \;\;\;\;Avg \\
            %\midrule \midrule
\shline
            {ED~\cite{RED}} & 43.8 & 40.9 & 38.7 & 36.8 & 34.6 & 33.9 & 32.5 & 31.6 & 36.6 \\
            {RED~\cite{RED}} & 45.3 & 42.1 & 39.6 & 37.5 & 35.8 & 34.4 & 33.2 & 32.1 & 37.5 \\
            {TRN~\cite{TRN}} & 45.1 & 42.4 & 40.7 & 39.1 & 37.7 & 36.4 & 35.3 & 34.3 & 38.9 \\
            {\bf{OadTR}} & 50.2 & 49.3 & 48.1 & 46.8 & 45.3 & 43.9 & 42.4 & 41.1 & \textbf{45.9} \\ \cdashline{1-10}
            {\bf{OadTR (Kinetics)}} & 59.8 & 58.5 & 56.6 & 54.6 & 52.6 & 50.5 & 48.6 & 46.8 & \textbf{53.5} \\
            % \bottomrule
        \end{tabular}
        \vspace{-3pt}
        \caption{
            Results on the THUMOS14 dataset in terms of mAP (\%).
        }        
        \vspace{3pt}
    \end{subtable}
    \vspace{-10pt}
    \caption{
        Action anticipation results of our OadTR compared to state-of-the-art
        methods using the same two-stream features.
    }
    \vspace{-2pt} 
    \label{table:anticipation}
\end{table*}
%
%
%
%
% pooling type ablation
%
\begin{table}[t]
    \centering
    \footnotesize
\tablestyle{4pt}{1.05}
% \small
    \begin{tabular}
        {@{\;\;}l@{\;\;}@{\;\;}|l@{\;\;}@{\;\;}c@{\;\;}@{\;\;}c@{\;\;}@{\;\;}c@{\;\;}}
        %\toprule
        & & \multicolumn{3}{c}{Settings}\\
        % \cmidrule(r){3-5}  
\cline{3-5}
        Method & Dataset & Max-pool & Avg-pool & w/o encoder  \\
        %\midrule
\shline
%\cmidrule(r){2-5}
        \multirow{3}{*}{OadTR}
        & {HDD}   & 29.2 & \bf{29.8} & 26.1 \\
%\cmidrule(r){2-5}
        %\midrule
       %
        & {TVSeries}   & 85.1 & \bf{85.4} & 80.8 \\
%\cmidrule(r){2-5}
%
        %\midrule
        %
        & {THUMOS14}   & 57.9 & \bf{58.3} & 53.5 
%
%        \bottomrule

    \end{tabular}
    \vspace{-5pt}
    \caption{
        Comparison of different fusion methods and the necessity of the encoder feature encoding.
    }
    \vspace{-3pt}
    \label{table:aggregation_type}
\end{table}
%
%
%
%
%
%
% %------------------------------------------------------------------------- 
% implemented
\noindent\textbf{Importance of position encoding.} To demonstrate the importance of using position encoding, we organized some comparative experiments. As shown in Table~\ref{table:Ablation_study} (d), position encoding is necessary, and the learnable position encoding achieves the best result.
\vspace{+3pt}
\\
\textbf{Effect of head number.} Multi-head self-attention is a critical component. Here we study the impact of different head numbers of the decoder on performance. We can find that a optimal result is achieved when the number of heads is 4 (Table~\ref{table:Ablation_study} (e)). 
\vspace{+3pt}
\\
\textbf{Effect of query dimensions.} We conduct experiments to examine how different query dimensions affect online action detection performance. As shown in Tabel~\ref{table:Ablation_study} (f), when the number of feature dimensions is relatively small (\eg, 128), the model capacity is limited and the performance is relatively poor. As the number of feature dimensions gradually increases, the model capacity increases, and performance improves. However, when it exceeds a specific value (\eg, 1024), over-fitting may occur. 
\vspace{+3pt}
\\
% sparse_Transformer in the decoder
\textbf{Generalizability.} To further investigate the generalization of our OadTR for many Transformer variants, we replace the standard Transformer~\cite{Transformer} with Sparse Transformer~\cite{sparse_Transformer}. As illustrated in Table~\ref{table:Ablation_study} (g), we can find that the performance is still good after replacement. Generally, Sparse Transformer can reduce computational consumption but leads to a little performance degradation.
\vspace{+3pt}
\\
\textbf{Effect of decoder step count.} The step size used to predict the future also has an impact on performance. In Table~\ref{table:decoder}, we compare four step sizes (\ie, 2, 4, 8, and 16), and the results denote that $steps=8$ achieves the best results on the three datasets. % The results denote that for the online action detection task, $steps=8$ achieves the best results on the three datasets.
\vspace{+3pt}
\\
% we conduct experiments to meticulously explore that
\textbf{Effect of encoding layers $N$ and decoding layers $M$.} To further study the impact of different encoding and decoding layers on performance, additional experiments are conducted, and results are shown in Table~\ref{table:encoder_decoder}. In most cases, the best results are achieved when $N=3, M=5$. However, there will also be fluctuations, such as on the TVSeries dataset. 
\vspace{+3pt}
\\
\textbf{Feature aggregation type.} We also conduct experiments to explore different types that aggregate future and current features. We can notice that Avg-pool is better than Max-pool (Table~\ref{table:aggregation_type}). The reason may be that the predicted deep semantic representations of different time steps all have a specific promotion effect on the current classification. Simultaneously, the results of \textit{w/o encoder} also indicate the necessity of the encoder to learn discriminative features. 

%
%
%
%
%-------------可视化------------------------------------------------------------
\begin{figure}[t!]
\centering
%\begin{minipage}[t]{0.499\linewidth}
{\centering{\includegraphics[width=.48\linewidth]{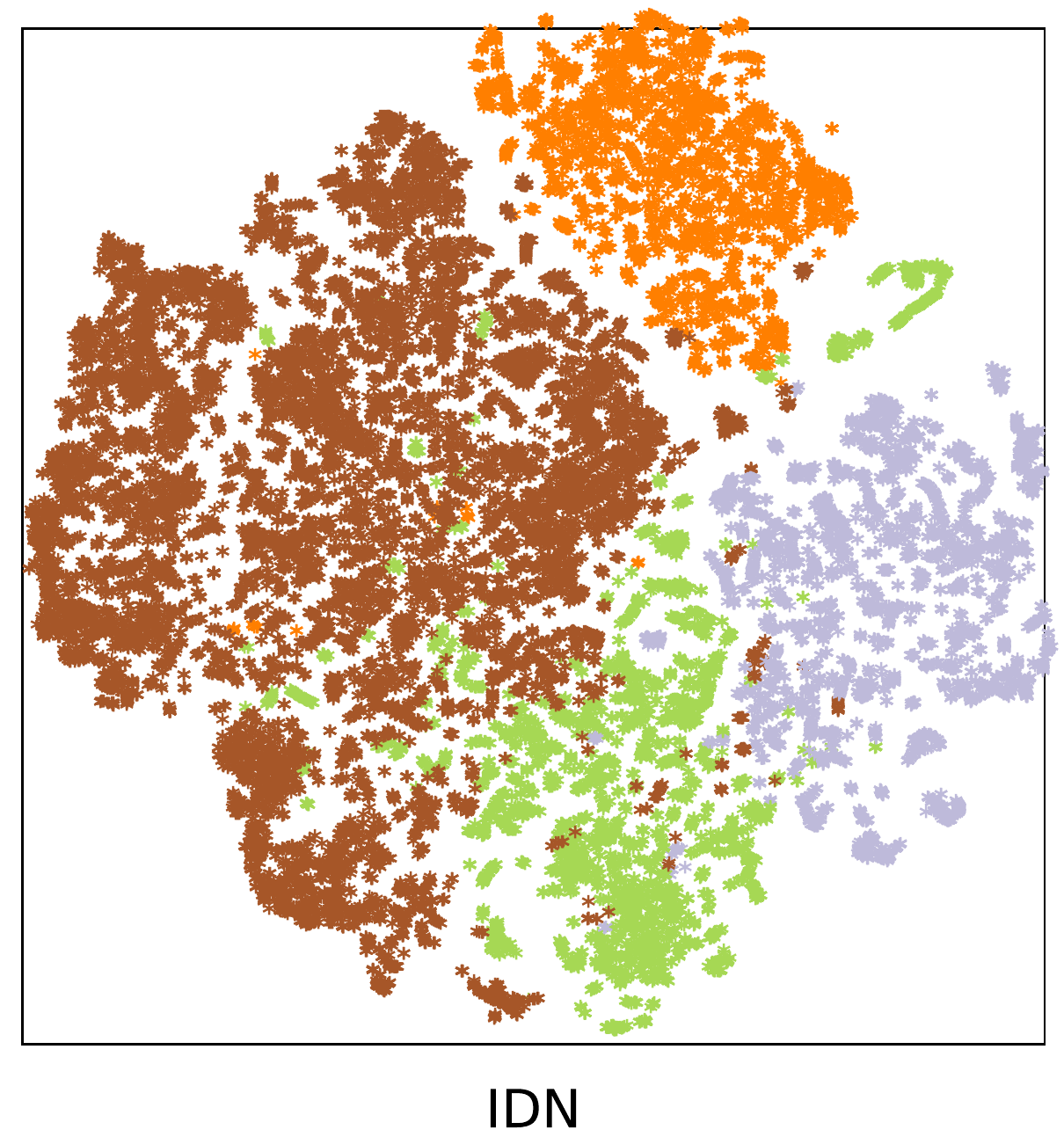}}}
% \centering{\small{(a) Information Discrimination Unit (IDU)}}
%\end{minipage}
\hspace{1mm}
%\begin{minipage}[t]{0.495\linewidth}
\centering{\includegraphics[width=.48\linewidth]{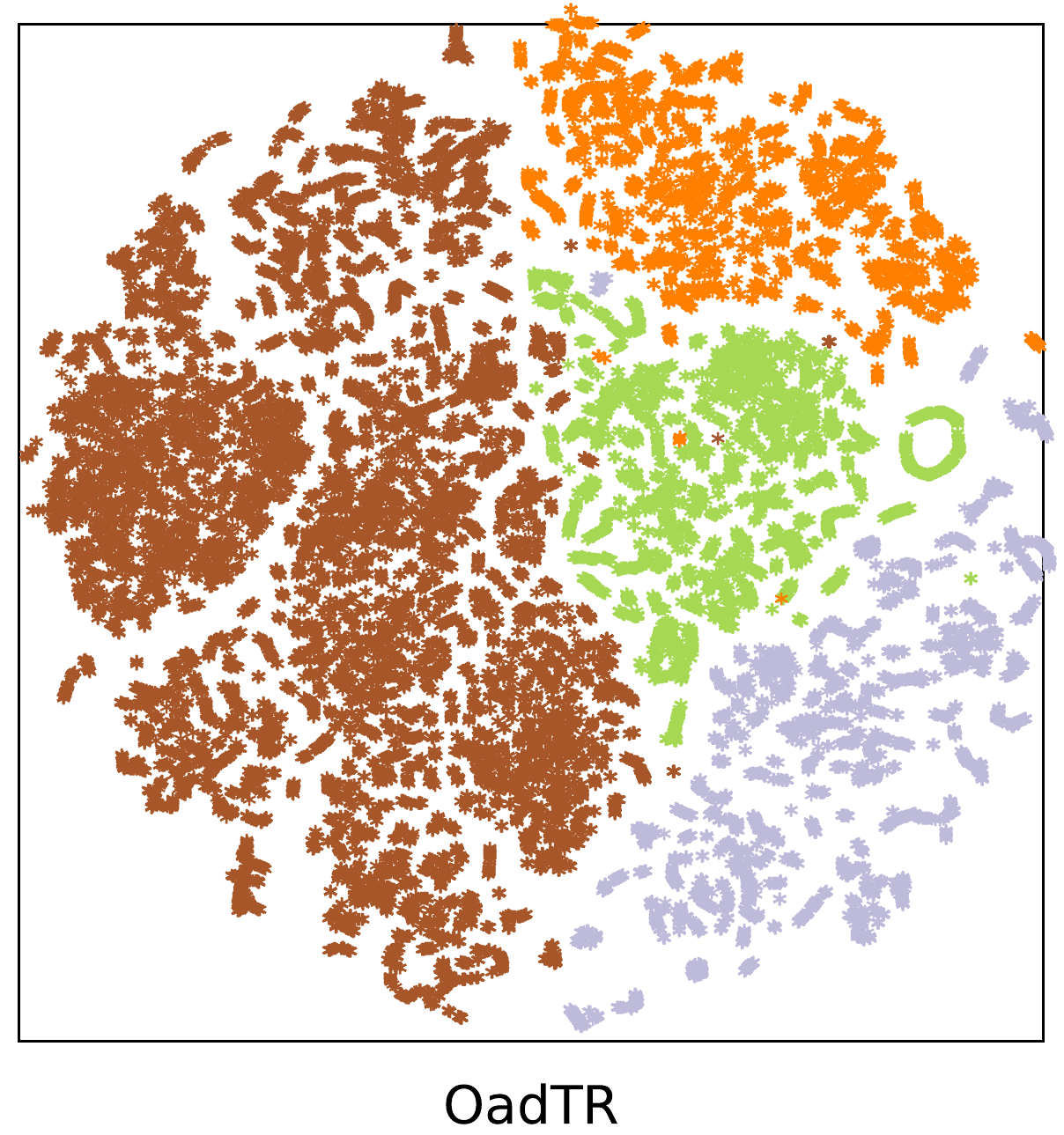}} 
% \centering{\small{(b) Information Discrimination Network (IDN)}}
%\end{minipage}
\vspace{-6mm}
\caption{\label{figure-visual-1}The t-SNE visualization of the classification embedding logits. Different colors correspond to different action categories from the THUMOS14 dataset. The mutual color correspondences include: \textcolor[RGB]{241,125,31}{CliffDiving}, \textcolor[RGB]{148,181,81}{HighJump}, \textcolor[RGB]{104,50,6}{PoleVault} and \textcolor[RGB]{211,222,252}{Shotput}. Better view in colored PDF.
% \textcolor{orange}{orange} $\Leftrightarrow $ CliffDiving; \textcolor{green}{cyan-blue} $\Leftrightarrow $ HighJump; \textcolor{brown}{brown} $\Leftrightarrow $ PoleVault; \textcolor{magenta}{purple} $\Leftrightarrow $ Shotput. Better view in colored PDF.
}
\vspace{-4mm}
\end{figure}
%
% \textcolor{red/blue/green/black/white/cyan/magenta/yellow}{text}
%
%
%
%
\begin{figure}[t!]
\centering
\centering{\includegraphics[width=.99\linewidth]{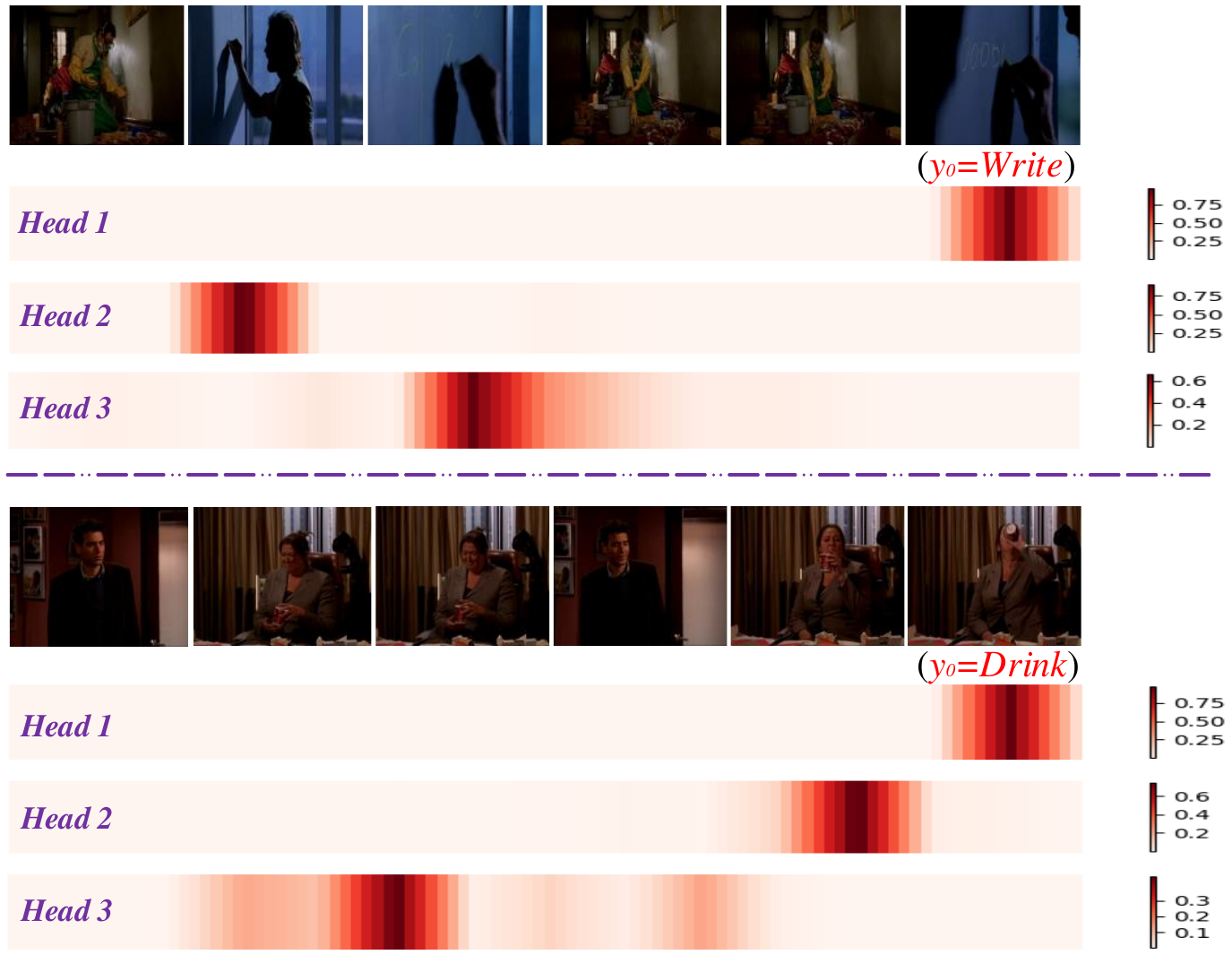}} 
% \centering{\small{(b) Information Discrimination Network (IDN)}}
%\end{minipage}
\vspace{-1mm}
\caption{\label{figure-visual-2}Attention visualization maps. They indicate how much attention is paid to parts of the input streaming video. 
}
\vspace{-4mm}
\end{figure}
\subsection{Action anticipation}
In our proposed OadTR, we introduce predicted future information to identify current actions. To evince the accuracy of our predictions, we also conduct experiments to compare with other methods. Table~\ref{table:anticipation} demonstrates that OadTR outperforms the current state-of-the-art method~\cite{TRN} by a large margin.
In particular, the performance of OadTR is 2.1$\%$ higher than TRN~\cite{TRN} on the TVSeries dataset and 7.0$\%$ higher on THUMOS14. Furthermore, the performance of OadTR can also be further improved when pre-training on Kinetics. % the performance of OadTR pre-training on Kinetics can also be further improved.  
\subsection{Qualitative evaluation}
% method (\ie, IDN~\cite{IDN})
For better analysis, we visualize the classification results in Figure~\ref{figure-visual-1}. Obviously, by visualizing all the test samples of the four action categories, we can observe that our OadTR has better separability compared with the current state-of-the-art IDN~\cite{IDN}.
Further, we show OadTR's multi-head attention visualization results in Figure~\ref{figure-visual-2}, which indicates the importance of the multi-head design to learn more complex and comprehensive relations between different neighboring frame chunks. 
%
%
%
%
%
%##################################################################################################

\section{Conclusion}
In this paper, we have proposed a new online action detection framework built upon Transformers, termed OadTR. 
In contrast to existing RNN based methods that process the sequence one by one recursively and hard to be optimized, we aim to design a direct end-to-end parallel network.
OadTR can recognize current actions by encoding historical information and predicting future context simultaneously.
Extensive experiments are conducted and verify the effectiveness of OadTR. Particularly, OadTR achieves higher training and inference speeds than current RNN based approaches, and obtains significantly better performance compared to the state-of-the-art methods.
In the future, we will extend our OadTR model to more tasks such as action recognition, spatio-temporal action detection, \etc.
%
%
%
%-------------------------------------------------------------------------

{\small
\bibliographystyle{ieee_fullname}
\bibliography{egbib}
}

\end{document}